\documentclass[11pt]{article}
\pdfoutput=1

\usepackage{fullpage}
\usepackage{amsfonts}
\usepackage{amssymb}
\usepackage{amsmath}
\usepackage{amsthm}
\usepackage{mathptm}
\usepackage{latexsym}
\usepackage{calc}
\usepackage{ifthen}
\usepackage{times} 
\usepackage{xspace}
\usepackage{graphicx}
\usepackage{wrapfig}
\usepackage{rotating}

\usepackage{color}
\definecolor{linkColor}{rgb}{0.1,0.1,0.8}
\usepackage[colorlinks=true,bookmarks,linkcolor=linkColor,citecolor=linkColor,pagecolor=blue,urlcolor=blue,pdftitle={Seshia paper},pdfauthor={Sanjit A. Seshia},plainpages=false]{hyperref}

\usepackage{program}
\usepackage{clrscode}
\usepackage{booktabs}

\definecolor{fixmeColor}{rgb}{1.0,0.2,0.2}

\newcommand{\comment}[1]{}
\newcommand{\hide}[1]{}

\newcounter{myctr}
\newenvironment{mylist}{\begin{list}{\arabic{myctr}.}
{\usecounter{myctr}
\setlength{\topsep}{1mm}\setlength{\itemsep}{0.25mm}
\setlength{\parsep}{0.1mm}
\setlength{\itemindent}{2mm}\setlength{\partopsep}{0mm}
\setlength{\labelwidth}{15mm}
\setlength{\leftmargin}{4mm}}}{\end{list}}

\newenvironment{myitemize}{\begin{list}{$\bullet$}
{\setlength{\topsep}{1mm}\setlength{\itemsep}{0.25mm}
\setlength{\parsep}{0.1mm}
\setlength{\itemindent}{0mm}\setlength{\partopsep}{0mm}
\setlength{\labelwidth}{15mm}
\setlength{\leftmargin}{4mm}}}{\end{list}}

\title{Towards Verified Artificial Intelligence}

\author{Sanjit A. Seshia$^*$, Dorsa Sadigh$^\dagger$, and S. Shankar Sastry$^*$\\
~~~\\
$^*$University of California, Berkeley ~~~~~ $^\dagger$ Stanford University\\
\{sseshia,sastry\}@eecs.berkeley.edu ~~~~~ dorsa@cs.stanford.edu}

\date{July 21, 2020}

\begin{document}

\maketitle

\begin{abstract}
{\em Verified artificial intelligence (AI)} is the goal of designing 
AI-based systems that have strong, ideally provable, assurances of 
correctness with respect to mathematically-specified requirements.
This paper considers Verified AI from a formal methods perspective.
We describe five challenges for achieving Verified AI, and 
five corresponding principles for addressing these challenges.
\end{abstract}

\section{Introduction}
\label{sec:intro}

{\em Artificial intelligence} (AI) is a term used for computational systems that attempt to mimic 
aspects of human intelligence, including functions we intuitively associate with human minds
such as `learning' and `problem solving' (e.g., see~\cite{nap-study-2017}).
Russell and Norvig~\cite{russell2010artificial} describe AI as
the study of general principles of
rational agents and components for constructing these agents.
We interpret the term AI broadly to include closely-related areas
such as machine learning (ML)~\cite{mitchell-97}.
Systems that heavily use AI, henceforth referred to as {\em AI-based
systems}, have had a significant impact in society in domains
that include healthcare, transportation, finance, social networking, 
e-commerce, education, etc. 
This growing societal-scale impact has brought with it a set of risks and concerns
including errors in AI software, cyber-attacks, and 
safety of AI-based systems~\cite{russell2015letter,dietterich2015rise,amodei2016concrete}.
Therefore, the question of verification and validation of AI-based
systems has begun to demand the attention of the research community. 
We define ``{\em Verified AI}'' as the goal of designing AI-based systems
that have strong, ideally provable, assurances of 
correctness with respect to mathematically-specified requirements.
How can we achieve this goal?

A natural starting point is to consider {\em formal methods} ---
a field of computer science and engineering
concerned with the rigorous mathematical specification, design, and
verification of systems~\cite{wing-ieeecomp90,clarke-acmcs96}. At its core, formal methods is
about {\em proof}: formulating specifications that form proof
obligations, designing systems to meet those obligations, and
verifying, via algorithmic proof search, that the systems indeed meet
their specifications. 
A spectrum of formal methods, from specification-driven testing and simulation~\cite{foster-abv09},
to model checking~\cite{clarke-lop81,queille-sympprog82,clarke-00} and 
theorem proving (see, e.g.~\cite{owre-pvs-cade92,kaufmann-00,gordon-93})  
are used routinely in the computer-aided design of integrated circuits and
have been widely applied to find bugs in software, analyze embedded systems,
and find security vulnerabilities.
At the heart of these advances are computational proof engines such as
Boolean satisfiability (SAT) solvers~\cite{malik-cacm09}, 
Boolean reasoning and manipulation routines based on Binary Decision Diagrams
(BDDs)~\cite{bryant-ieeetc86}, and
satisfiability modulo theories (SMT) solvers~\cite{barrett-smtbookch09}.

In this paper, we consider the challenge of Verified AI from a formal
methods perspective. That is, we review the manner in which formal methods
have traditionally been applied, 
analyze the challenges this approach may face for AI-based systems,
and propose ideas to overcome these challenges.
We emphasize that our discussion is focused on the role of formal methods
and does not cover the broader set of techniques that could be used to
improve assurance in AI-based systems.
Additionally, we seek to identify challenges applicable to a broad range
of AI/ML systems, and not limited to specific technologies such as deep neural
networks (DNNs) or reinforcement learning (RL) systems.
Our view of the challenges is largely shaped by problems arising from
the use of AI and ML in autonomous and semi-autonomous systems, though 
we believe the ideas presented here apply more broadly.

We begin in Sec.~\ref{sec:background} with some brief background on formal
verification and an illustrative example. We then outline challenges for Verified AI in
Sec.~\ref{sec:challenges} below, and describe ideas to address
each of these challenges in Sec.~\ref{sec:principles}.\footnote{The first
version of this paper was published in July 2016 in response to the
call for white papers for the CMU
Exploratory Workshop on Safety and Control for AI held in June 2016, and a second
version in October 2017.
This is the latest version reflecting the evolution of the authors' view 
of the challenges and approaches for Verified AI.}

\section{Background and Illustrative Example}
\label{sec:background}

\begin{figure}[h]
\centering
\includegraphics[width=0.7\columnwidth]{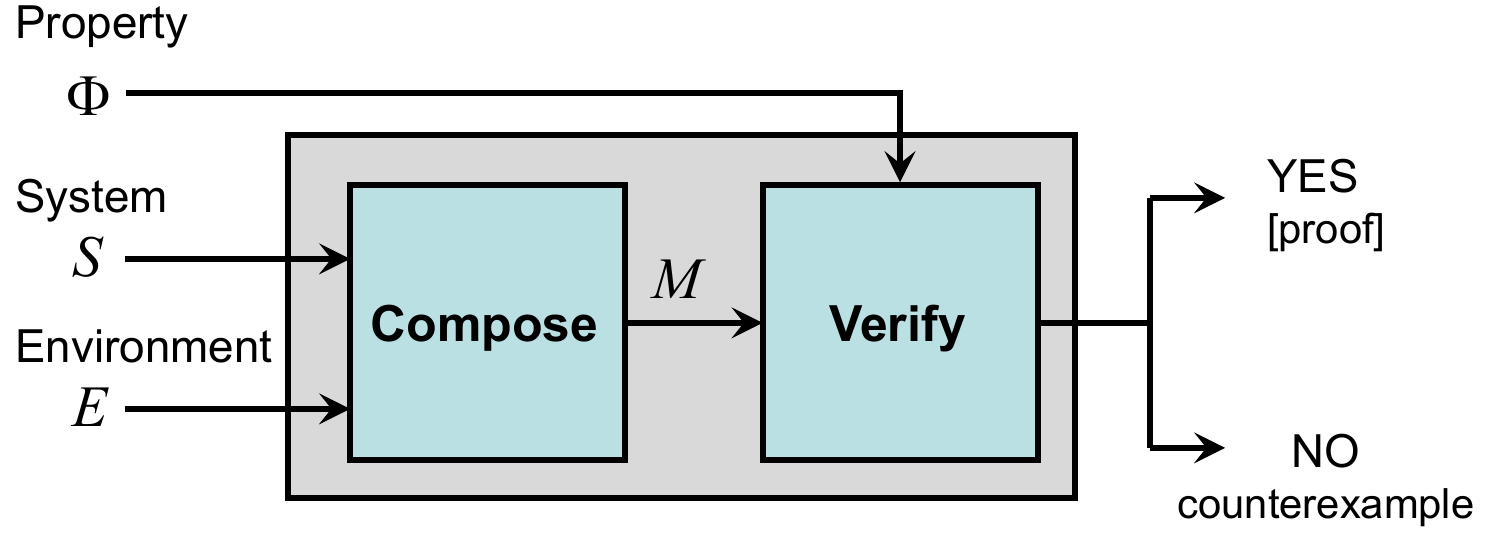}
\caption{\label{fig:formal-verif-flow}
Formal verification procedure.}
\end{figure}

Consider the typical formal verification process
as shown in Figure~\ref{fig:formal-verif-flow},
which begins with the following three inputs:
\begin{mylist}
\item A model of the system to be verified, $S$;
\item A model of the environment, $E$, and
\item The property to be verified, $\Phi$.
\end{mylist}
The verifier generates as output a YES/NO answer, indicating whether or not
$S$ satisfies the property $\Phi$ in environment $E$. Typically, a NO output
is accompanied by a counterexample, also called an error trace, which is an execution
of the system that indicates how $\Phi$ is violated. 
Some formal verification tools also include a proof or certificate
of correctness with a YES answer.
In this paper, we take a broad view of formal methods: any technique
that uses some aspect of formal specification, or verification, or synthesis,
is included. 
For instance, we include simulation-based hardware verification methods
or model-based testing methods for software since they use formal specifications
or models to guide the process of simulation or testing.

In order to apply formal verification to AI-based systems, at a
minimum, one must be able to represent the three inputs $S$, $E$ and
$\Phi$ in formalisms for which (ideally) there exist efficient decision
procedures to answer the YES/NO question as described above.
However, as we describe in Sec.~\ref{sec:challenges},
even constructing good representations of the three inputs
is not straightforward, let alone dealing with the complexity of
the underlying decision problems and associated design issues.

\begin{figure}[h]
\centering
\includegraphics[width=0.7\columnwidth]{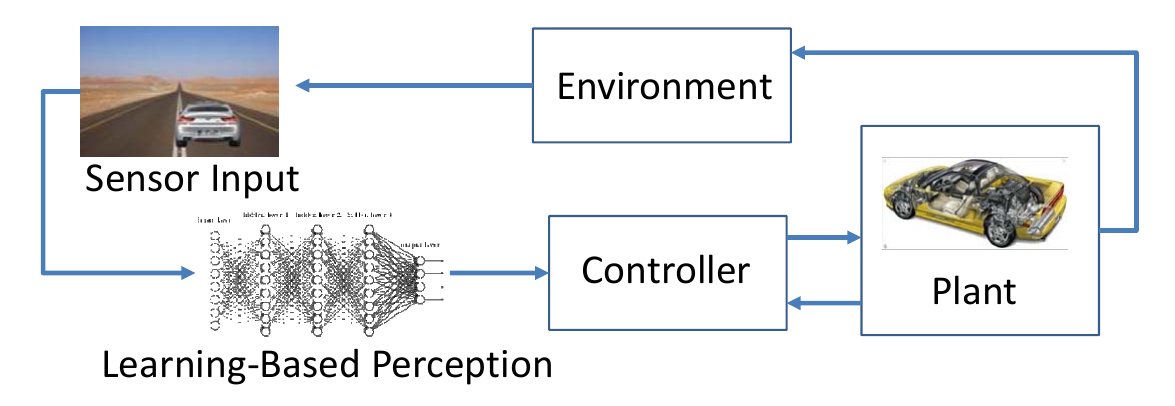}
\caption{\label{fig:cpsml-example} Example of closed-loop cyber-physical system with machine learning components (introduced in~\cite{dreossi-nfm17}).}
\end{figure}

We will illustrate the ideas in this paper with examples from the domain
of (semi-)autonomous driving. Fig~\ref{fig:cpsml-example} shows 
an illustrative example of an AI-based system: 
a closed-loop cyber-physical system comprising a semi-autonomous vehicle
with machine learning components along with its environment. 
Specifically, assume that the semi-autonomous ``ego vehicle'' 
has an automated emergency braking system (AEBS) 
that attempts to detect and classify objects in front of it and actuate
the brakes when needed to avert a collision.
Figure~\ref{fig:cpsml-example} shows the AEBS as a system
composed of a controller (automatic braking), a plant (vehicle sub-system under control including other parts of the autonomy stack), and a sensor (camera) along
with a perception component implemented using a deep neural network.  
The AEBS, when combined with the vehicle's environment, forms a closed loop cyber-physical system.
The controller regulates the acceleration and braking of the plant using the velocity of
the ego vehicle and the distance between it and an obstacle.
The environment of the ego vehicle comprises both agents and objects outside the vehicle (other vehicles, pedestrians, road objects, etc.) as well as inside the vehicle (e.g., a driver).
A safety requirement for this closed loop system can be informally characterized as the property of maintaining a safe distance between the moving ego vehicle and any other agent or object on the road.
However, as we will see in Sec.~\ref{sec:challenges}, there are many nuances to the specification, modeling, and verification of a system such as this one.

\section{Challenges for Verified AI} 
\label{sec:challenges}

We identify five major challenges to achieving formally-verified AI-based
systems, described in more detail below. 

\subsection{Environment Modeling}
\label{sec:env-challenge}

The {\em environments} in which AI/ML-based systems operate
can be very complex, with considerable uncertainty even about how many
and which agents are in the environment (both human and robotic), 
let alone about their intentions
and behaviors.
As an example, consider the difficulty in
modeling urban traffic environments
in which an autonomous car must operate.
Indeed, AI/ML is often introduced into these systems
precisely to deal with such complexity and uncertainty!
From a formal methods perspective, this makes it very hard to
create realistic environment models with respect to which
one can perform verification or synthesis.

We see the main challenges for environment modeling as being threefold:
\begin{myitemize}
\item {\em Unknown Variables:}
In the traditional success stories for formal verification, such as
verifying cache coherence protocols or device drivers, the {\em interface}
between the system $S$ and its environment $E$ is well-defined. 
The environment can only influence the system through this interface.
However, for AI-based systems, such as an autonomous vehicle 
example of Sec.~\ref{sec:background}, it may
be impossible to precisely define all the variables (features) of
the environment. Even in restricted scenarios where the environment
variables (agents) are known, there is a striking lack of information,
especially at design time, about their behaviors. Additionally,
modeling sensors such as LiDAR that represent the interface to the 
environment is in itself a major technical challenge.

\item {\em Modeling with the Right Fidelity:}
In traditional uses of formal verification, it is usually
acceptable to model the environment as a {\em non-deterministic} process subject to
constraints specified in a suitable logic or automata-based
formalism. Typically such an environment model is termed as being 
``over-approximate'', meaning that it may include (many) more environment
behaviors than are possible. Over-approximate environment modeling permits
one to perform sound verification without a detailed environment
model, which can be inefficient to reason with and hard to obtain.
However, for AI-based autonomy, purely non-deterministic modeling is 
likely to produce highly over-approximate models, which in turn
yields too many spurious bug reports, rendering
the verification process useless in practice. Moreover, many AI-based systems
make distributional assumptions on the environment, thus requiring
the need for probabilistic modeling; however, it can be difficult to
exactly ascertain the underlying distributions. One can address this by
learning a probabilistic model from data, but in this case it is important
to remember that the model parameters
(e.g., transition probabilities) are only estimates, not
precise representations of environment behavior.
Thus, verification algorithms cannot consider the resulting probabilistic 
model to be ``perfect''; we need to represent uncertainty in the
model itself.

\item {\em Modeling Human Behavior:}
For many AI-based systems, such as semi-autonomous
vehicles, human agents are a key part of the environment and/or system.
Researchers have attempted modeling humans as non-deterministic or 
stochastic processes with the goal of verifying the
correctness of the overall system~\cite{rushby2002using,sadigh-aaaifv14}. 
However, such approaches must deal with the variability and uncertainty in human behavior.
One could take a data-driven approach based on machine learning (e.g.,~\cite{ng-icml00}), but
such an approach is sensitive to the expressivity of the features
used by the ML model and the quality of data.
In order to achieve Verified AI for such human-in-the-loop systems, we need to address the
limitations of current human modeling techniques and provide
guarantees about their {\em prediction accuracy and convergence}. 
When learned models are used, one must represent any {\em uncertainty 
in the learned parameters} as a first-class entity in the model,
and take that into account in verification and control.

\end{myitemize}
The first challenge, then, is to come up with a systematic
method of environment
modeling that allows one to provide provable guarantees on the
system's behavior even when there is considerable uncertainty about 
the environment.

\subsection{Formal Specification}
\label{sec:spec-challenge}

Formal verification critically relies on having a formal specification
--- a precise, mathematical statement of what the system is supposed to
do. However, the challenge of coming up with a high-quality formal
specification is well known, even in application domains in which
formal verification has found considerable success (see, e.g.,~\cite{beer-fmsd01}).
This challenge is only exacerbated in AI-based systems. We identify
three major problems.

\noindent
{\em Specification for Hard-to-Formalize Tasks:} 
Consider the perception module in the AEBS controller of Fig.~\ref{fig:cpsml-example}
which must detect and classify objects, distinguishing vehicles and pedestrians from 
other objects. Correctness for this module in the classic formal methods sense 
requires a formal definition of each type of road user, which is extremely difficult, if not impossible.
Similar problems arise for other tasks involving perception and communication, such as natural language processing.
How then, do we specify correctness properties for such a module? 
What should the specification language be and what
tools can one use to construct a specification? 

\noindent
{\em Quantitative vs. Boolean Specifications:}
Traditionally, formal specifications tend to be Boolean, mapping a
given system behavior to true or false. However, in AI and ML, 
specifications are often given as objective functions specifying
costs or rewards. Moreover, there can be multiple objectives, some of
which must be satisfied together, and others that may need to be traded off against each other
in certain environments.
What are the best ways to unify Boolean and quantitative approaches to specification?
Are there formalisms that can capture commonly discussed properties of AI components
such as robustness and fairness in a unified manner?

\noindent
{\em Data vs. Formal Requirements:}
The view of ``data as specification'' is common in machine learning.
Labeled ``ground truth'' data is often the only specification of correct behavior.
On the other hand, a specification in formal methods is a 
mathematical property that defines the set of correct behaviors. 
How can we bridge this gap?

Thus, the second challenge is to design effective methods to specify desired
and undesired properties of systems that use AI- or ML-based components.

\subsection{Modeling Learning Systems}
\label{sec:sys-challenge}

In most traditional applications of formal verification, the system
$S$ is precisely known: it is a program or a circuit described in a
programming language or hardware description language. 
The system modeling problem is primarily concerned with reducing the
size of $S$ to a more tractable one by abstracting away irrelevant details. 

AI-based systems lead to a very different challenge for system
modeling, primarily stemming from the use of
{\em machine learning}: 
\begin{myitemize}

\item 
{\em Very high-dimensional input space:}
ML components used for perception usually operate over very
high-dimensional input spaces. For the illustrative example
of Sec.~\ref{sec:background} from~\cite{dreossi-nfm17}, each
input RGB image is of dimension $1000\times600$ pixels, 
contains $256^{1000\times 600 \times 3}$ elements, and in general
the input is a stream of such high-dimensional vectors.
Although formal methods has been used for high-dimensional
input spaces (e.g., in digital circuits), the nature of the input
spaces for ML-based perception is different -- not entirely Boolean,
but hybrid, including both discrete and continuous variables.

\item
{\em Very high-dimensional parameter/state space:}
ML components such as deep neural networks have anywhere from
thousands to millions of model parameters and primitive components. 
For example, state-of-the-art DNNs used by the authors in instantiations of the
example of Fig.~\ref{fig:cpsml-example} have up to
60 million parameters and tens of layers.
This gives rise to a huge search space for verification that requires
careful abstraction.

\item
{\em Online adaptation and evolution:}
Some learning systems, such as a robot using reinforcement learning,
evolve as they encounter new data and situations. For such systems,
design-time verification must either account for future changes in the behavior
of the system, or else be performed incrementally and online as
the learning system evolves.

\item
{\em Modeling systems in context:}
For many AI/ML components, their specification is only defined by
the context. For example, verifying robustness of a DNN such as
the one in Fig.~\ref{fig:cpsml-example} requires us to capture
a model of the surrounding system.
We need techniques to model
ML components along with their {\em context} so that semantically
meaningful properties can be verified.

\end{myitemize}

\subsection{Efficient and Scalable  Design and Verification of Models and Data} 
\label{sec:train-challenge}

The effectiveness of formal methods in the domains of hardware and
software has been driven by advances in underlying ``computational
engines'' --- e.g., SAT, SMT, numerical simulation, and model checking.
Given the scale of AI/ML systems, the complexity of their environments,
and the new types of specifications involved, several advances are 
needed in creating computational engines for efficient and
scalable training, testing, design, and verification of AI-based systems.
We identify here the key challenges that must be overcome
in order to achieve these advances.

\noindent
{\em Data Generation:}
Data is the fundamental starting point for machine learning.
Any quest to improve the quality of a machine learning system
must improve the quality of the data it learns from. 
Can formal methods help to systematically select, design and augment
the data used for machine learning?

We believe the answer is yes, but that more needs to be done.
Formal methods has proved effective for the systematic generation of
counterexamples and test data that satisfy constraints including for 
simulation-based verification of circuits (e.g.,~\cite{kitchen-iccad07}) 
and finding security exploits in commodity software (e.g.,~\cite{brumley-cacm14}).
However, the requirements for AI/ML systems are different.
The types of constraints can be much more complex, e.g., encoding
requirements on ``realism'' of data captured using sensors from
a complex environment such as a traffic situation.
We need to generate not just single data items, but
an ensemble that satisfies distributional constraints.
Additionally, data generation must be selective, e.g., meeting objectives
on data set size and diversity for effective training and generalization.
All of these additional requirements necessitate the development of
a new suite of formal techniques.

\hide{
AI-based systems benefit from testing not just by gaining a higher level of
assurance, but also by leveraging the generated data for retraining.
Moreover, recent efforts have shown that various machine learning algorithms can
fail under small adversarial perturbations (e.g.,~\cite{nguyen2015deep,fawzi2015analysis,moosavi2015deepfool,goodfellow2014explaining,papernot-eurosp16}).
Learning algorithms promise to generalize from data, but such simple
perturbations that fool the algorithms create concerns
regarding their use in safety-critical applications such as autonomous
driving.
Such small perturbations might be even unrecognizable to humans, but
drive the algorithm to misclassify the perturbed data. 
Further, we need to generate not just single data items, but
an ensemble that is ``realistic'' and satisfies distributional
constraints.

Thus, the question is: can we devise techniques based on
formal methods to systematically generate training and testing data
for ML-based components?
}

\noindent
{\em Quantitative Verification:}
Several safety-critical applications of AI-based systems are in
robotics and cyber-physical systems. In such systems, the
scalability challenge for verification can be very high.
In addition to the scale of systems as measured by traditional metrics
(dimension of state space, number of components, etc.), the {\em
types} of components can be much more complex. For instance, in
(semi-)autonomous driving, autonomous vehicles and their controllers
need to be modeled as {\em hybrid systems} combining both discrete and
continuous dynamics. 
Moreover, agents in the environment (humans, other vehicles) may need
to be modeled as {\em probabilistic processes}.
Finally, the requirements may involve not only traditional Boolean
specifications on safety and liveness, but also {\em quantitative
requirements} on system robustness and performance. 
Yet, most of the existing verification methods are targeted towards
answering Boolean verification questions.
To address this gap, new scalable engines for quantitative 
verification must be developed.

\noindent
{\em Compositional Reasoning:}
In order for formal methods to scale to large AI/ML systems, 
compositional (modular) reasoning is essential. In compositional
verification, a large system (e.g., program) is split up into its
components (e.g., procedures), each component is verified against
a specification, and then the component specifications together 
entail the system-level specification. A common approach for
compositional verification is the use of {\em assume-guarantee contracts}. For
example, a procedure assumes something about its starting state
(pre-condition)
and in turn guarantees something about its ending state (post-condition).
Similar assume-guarantee paradigms have been developed for concurrent
software and hardware systems.
A theory of assume-guarantee contracts does not yet exist for AI-based systems.

Moreover, AI/ML systems pose a particularly vexing challenge for compositional
reasoning. Compositional verification requires
{\em compositional specification} --- i.e.,
the components must be formally-specifiable.
However, as noted in Sec.~\ref{sec:spec-challenge}, it may be impossible
to formally specify the correct behavior of a perception component.
One of the challenges, then, is to develop techniques for compositional
reasoning that do not rely on having complete compositional specifications~\cite{seshia-tr17}.
Additionally, more work needs to be done for extending the theory 
and application of
compositional reasoning to probabilistic systems and specifications.

\subsection{Correct-by-Construction Intelligent Systems}
\label{sec:scale-challenge}
\label{sec:correct-by-construction}

In an ideal world, verification should be integrated with
the design process so that the system is ``correct-by-construction.''
Such an approach could either interleave verification steps with
compilation/synthesis steps, such as in the register-transfer-level (RTL)
design flow common in integrated circuits, or devise synthesis
algorithms so as to ensure that the implementation
satisfies the specification, such as in reactive synthesis from
temporal logic~\cite{pnueli-popl89}.
Can we devise a suitable correct-by-construction design flow
for AI-based systems?

\noindent
{\em Specification-Driven Design of ML Components:}
Can we design, from scratch, a machine learning component (model)
that provably satisfies a formal specification?
(This assumes, of course, that we solve the formal
specification challenge described above in Sec.~\ref{sec:spec-challenge}.)
The clean-slate design of an ML component has many aspects:
(1) designing the data set,
(2) synthesizing the structure of the model,
(3) generating a good set of features,
(4) synthesizing hyper-parameters and other aspects of ML algorithm
selection, and
(5) automated techniques for debugging ML models or the
specification when synthesis fails.
More progress is needed on all these fronts.

\noindent
{\em Theories of Compositional Design}:
Another challenge is to design the overall system comprising
multiple learning and non-learning components. While theories of
compositional design have been developed for digital circuits
and embedded systems (e.g.~\cite{sangiovanni-ejc12,sifakis-pieee15}), 
we do not as yet have such theories for AI-based systems.
For example, if two ML models are used for perception on two different
types of sensor data (e.g., LiDAR and visual images), and individually
satisfy their specifications under certain assumptions, under what 
conditions can they be used together to improve the reliability of
the overall system?
And how can one design a planning component so as to overcome
limitations of an ML-based perception component that it receives
input from?

\noindent
{\em Bridging Design Time and Run Time for Resilient AI:}
Due to the complexity of AI-based systems and the environments in
which they operate, even if all the challenges for specification
and verification are solved, it is likely that one will not be able to prove
unconditional safe and correct operation. There will always be
situations in which we do not have a provable guarantee of correctness.
Therefore, techniques for achieving fault tolerance and error resilience
at run time must play a crucial role. In particular,
there is not yet a systematic understanding of
what can be achieved at design time, how the design process
can contribute to safe and correct operation of the AI 
system at run time, and how the design-time and run-time techniques
can interoperate effectively.

\section{Principles for Verified AI}
\label{sec:principles}

For each of the challenges described in the preceding
section, we suggest a corresponding set of
principles to follow in the design/verification
process to address that challenge. 
These five principles are:
\begin{mylist}
\item Use an {\em introspective, data-driven, and probabilistic} approach
to model the environment;
\item Combine formal specifications of {\em end-to-end behavior} with
{\em hybrid Boolean-quantitative formalisms} for learning systems and perception 
components and use {\em specification mining} to bridge the data-property gap;
\item For ML components, develop new {\em abstractions}, {\em explanations}, and {\em semantic analysis} techniques;
\item Create a new class of {\em compositional, randomized, and quantitative formal methods} for data generation, testing, and verification, and 
\item Develop techniques for {\em formal inductive synthesis} of AI-based systems and design of {\em safe learning systems}, supported by techniques for {\em run-time assurance}.
\end{mylist}
We have successfully applied these principles over the past few years,
and, based on this experience, believe that they provide a good starting
point for applying formal methods to AI-based systems.
Our formal methods perspective on the problem complements
other perspectives that have been expressed (e.g.,~\cite{amodei2016concrete}). 
Experience over the past few years provides evidence that the principles 
we suggest can point a way towards the goal of Verified AI.

\subsection{Environment Modeling: Introspection, Probabilities, and Data}

Recall from Sec.~\ref{sec:env-challenge}, the three challenges
for modeling the environment $E$ of an AI-based system $S$: 
{\it unknown variables},
{\it model fidelity}, and
{\it human modeling}.
We propose to tackle these challenges with three corresponding principles.

\noindent
{\em Introspective Environment Modeling:} 
We suggest to address the unknown variables problem by
developing design and verification methods that are {\em
introspective}, i.e., they algorithmically identify assumptions 
$A$ that system $S$
makes about the environment $E$ that are sufficient to guarantee the satisfaction of the specification $\Phi$~\cite{seshia-rv19}. The assumptions $A$ must be
ideally the {\em weakest} such assumptions, and also must be
{\em efficient to generate at design time and monitor at run time} 
over available sensors
and other sources of information about the environment
so that mitigating actions can be taken when they are violated. 
Moreover, if there is a human operator
involved, one might want $A$ to be translatable into an explanation
that is {\em human understandable}, so that $S$ can ``explain'' to the human
why it may not be able to satisfy the specification $\Phi$.
Dealing with these multiple requirements, as well as the need for good
sensor models, makes introspective
environment modeling a highly non-trivial task that requires
substantial progress~\cite{seshia-rv19}.
Preliminary work by the authors has shown that such extraction 
of monitorable assumptions is feasible in
very simple cases~\cite{li-tacas14}, although
more research is required to make this practical.

\noindent
{\em Active Data-Driven Modeling:} 
We believe human modeling requires an {\em active data-driven} approach.
Relevant theories from cognitive science and psychology, such as
that of bounded rationality~\cite{simon-90,russell-ai97},
must be leveraged, but it is important for those models to be
expressed in formalisms compatible with formal methods.
Additionally, while using a data-driven approach to infer a model,
one must be careful to craft the right model structure and features.
A critical aspect of human modeling is to capture {\em human intent}.
We believe a three-pronged approach is required: first, define
model templates/features based on expert knowledge; then,
use offline learning to complete the model for design time use,
and finally, 
{\em learn and update} environment models at run time by
monitoring and interact with the environment.
Initial work has shown how data gathered from driving simulators via human subject experiments 
can be used to generate models of human driver behavior that are useful
for verification and control of autonomous vehicles~\cite{sadigh-aaaifv14,sadigh-iros16}.

\noindent
{\em Probabilistic Formal Modeling:} In order to tackle the model fidelity challenge,
we suggest to use formalisms that combine probabilistic and non-deterministic modeling. 
Where probability distributions can be reliably specified or estimated, one can use
probabilistic modeling. In other cases, non-deterministic modeling can be used to over-approximate
environment behaviors. While formalisms such as Markov Decision Processes (MDPs) already
provide a way to blend probability and non-determinism, we believe techniques that blend probability
and logical or automata-theoretic formalisms, such as the paradigm of {\em probabilistic
programming}~\cite{blog,fremont-pldi19}, can provide an expressive and programmatic way to model environments.
We expect that in many cases, such probabilistic programs will need to be learned/synthesized (in part) from
data. In this case, any uncertainty in learned parameters
must be propagated to the rest of the system and represented in the
probabilistic model.
For example, the formalism of {\em convex Markov decision processes}
(convex MDPs)~\cite{Nilim05,puggelli-cav13,sadigh-aaaifv14} provide a way of
representing uncertainty in the values of learned transition
probabilities. Algorithms for verification and control may then need
to be extended to handle these new abstractions (see, e.g.,~\cite{puggelli-cav13}).

\subsection{End-to-End Specifications, Hybrid Specifications, and Specification Mining}

Writing formal specifications for AI/ML components is hard, arguably 
even impossible if the component imitates a human perceptual task.
Even so, we think the challenges described in Sec.~\ref{sec:spec-challenge}
can be addressed by following three guiding principles.

\noindent
{\em End-to-End/System-Level Specifications:}
In order to address the specification-for-perception challenge, 
let us change the problem slightly. We suggest to first focus on
{\em precisely specifying the end-to-end behavior} of the {\em AI-based system}.
By ``end-to-end'' we mean the specification on the entire closed-loop system 
(see Fig.~\ref{fig:cpsml-example}) or a precisely-specifiable sub-system containing the
AI/ML component, not on the component alone. Such a specification
is also referred to as a ``system-level'' specification.
For our AEBS example, this involves specifying the property $\Phi$
corresponding to maintaining a minimum distance from any object during motion.
Starting with such a system-level (end-to-end) specification, we then
derive from it constraints on the input-output interface of the
perception component that guarantee that the system-level specification
is satisfied. Such constraints serve as a partial specification under which
the perception component can be analyzed (see~\cite{dreossi-nfm17}).
Note that these constraints need not be human-readable.

\noindent
{\em Hybrid Quantitative-Boolean Specifications:} 
Boolean and quantitative specifications both have their advantages.
On the one hand, Boolean specifications are easier to compose.
On the other hand, objective functions lend themselves to optimization
based techniques for verification and synthesis, and to defining
finer granularities of property satisfaction.
One approach to bridge this gap is to move to quantitative specification languages,
such as logics with both Boolean and quantitative semantics (e.g. STL~\cite{maler-formats04})
or notions of weighted automata (e.g.~\cite{chatterjee-tocl10}).
Another approach is to combine both Boolean and quantitative specifications
into a common specification structure, such as a {\em rulebook}~\cite{censi-icra19},
where specifications can be organized in a hierarchy, compared,
and aggregated.
Additionally, novel formalisms bridging ideas from formal methods and machine learning
are being developed to model the different variants of properties such as robustness, fairness,
and privacy, including notions of semantic robustness (see, e.g.,~\cite{seshia-atva18,dreossi-vnn19}).

\noindent
{\em Specification Mining:}
In order to bridge the gap between data and formal specifications,
we suggest the use of techniques for inferring specifications from
behaviors and other artifacts --- so-called {\em specification mining} techiques
(e.g.,~\cite{ernst-phd-thesis,li-phd14}).
Such methods could be used for ML components in general, including
for perception components, since in many cases it is not required
to have an exact specification or one that is human-readable.
Specification mining methods could also be used to infer human intent
and other properties from demonstrations~\cite{vazquez-neurips18} or more complex forms of
interaction between multiple agents, both human and robotic.

\subsection{System Modeling: Abstractions, Explanations, and Semantic Feature Spaces}

Let us now consider the challenges, described in Sec.~\ref{sec:sys-challenge}, 
arising in modeling systems $S$ that learn from experience. 
In our opinion, advances in three areas are needed in order to address
these challenges:

\noindent
{\em Automated Abstraction:} 
Techniques for automatically generating abstractions of systems have
been the linchpins of formal methods, playing crucial roles in extending
the reach of formal methods to large hardware and software systems.
In order to address the challenges of very high dimensional hybrid
state spaces and input spaces for ML-based systems, 
we need to develop effective techniques to abstract ML models 
into simpler models that are more amenable to formal analysis.
Some promising advances in this regard include the use of abstract
interpretation to analyze deep neural networks (e.g.~\cite{gehr-ieeesp18}),
the use of abstractions for falsifying cyber-physical systems with
ML components~\cite{dreossi-nfm17}, 
and the development of probabilistic logics that capture guarantees 
provided by ML algorithms (e.g.,~\cite{sadigh-rss16-uncertainty}). 

\noindent
{\em Explanation Generation:}
The task of modeling a learning system can be made easier if the
learner accompanies its predictions with {\em explanations} of how
those predictions result from the data and background knowledge.
In fact, this idea is not new -- it has long been investigated by the
ML community under terms such as {\em explanation-based
generalization}~\cite{mitchell-ml86}. Recently, there has been
a renewal of interest in using logic to explain the output of
learning systems (e.g.~\cite{vazquez-cav17,jha-jar19}). Such approaches
to generating explanations that are
compatible with the modeling languages used in formal methods
can make the task of system modeling for verification considerably easier. 
ML techniques that incorporate causal and counterfactual reasoning~\cite{pearl-cacm19}
can also ease the generation of explanations for use with formal methods.

\noindent
{\em Semantic Feature Spaces:}
The verification and adversarial analysis~\cite{goodfellow-cacm18} of ML models is more meaningful
when the generated adversarial inputs and counterexamples
have semantic meaning in the context in which the ML models are used.
There is thus a need for techniques that can analyze ML models in the context of
the systems within which they are used, i.e., for {\em semantic adversarial analysis}~\cite{dreossi-cav18}.
A key step is to represent the {\em semantic feature space} modeling the
environment in which the ML system operates, as opposed to the concrete feature
space which defines the input space for the ML model. 
This follows the intuition that the semantically meaningful part of the
concrete feature space (e.g. images of traffic scenes) form a much lower
dimensional latent space as compared to the full concrete feature space.
For our illustrative example in Fig.~\ref{fig:cpsml-example},
the semantic feature space is the lower-dimensional space representing the 3D
world around the autonomous vehicle, whereas the concrete feature space is
the high-dimensional pixel space. Since the semantic feature space is lower
dimensional, it can be easier to search over (e.g.~\cite{dreossi-nfm17,huang-cav17}).
However, one typically needs to have a ``renderer'' that maps a point in the
semantic feature space to one in the concrete feature space, and certain 
properties of this renderer, such as differentiability~\cite{li-siggraphasia18}, make it easier to
apply formal methods to perform goal-directed search of the semantic feature space.

\subsection{Compositional and Quantitative Methods for Design and Verification of Models and Data}

Consider the challenge, described in Sec.~\ref{sec:train-challenge},
of devising computational engines for scalable training, testing,
and verification of AI-based systems. We see three promising
directions to tackle this challenge.

\noindent
{\em Controlled Randomization in Formal Methods:} 
Consider the problem of {\em data set design} -- i.e., systematically 
generating training data for a ML component in an AI-based system. 
This synthetic data generation problem has many facets. First, one must 
define the space of ``legal'' inputs so that the examples are 
well formed according to the application semantics.
Secondly, one might want to impose constraints on ``realism'', e.g.,
a measure of similarity with real-world data.
Third, one might need to impose constraints on
the distribution of the generated examples in order to obtain guarantees
about convergence of the learning algorithm to the true concept. 
What can formal methods offer towards solving this problem?

We believe that the answer may lie in a new class of {\em randomized formal methods}
-- randomized algorithms for generating test inputs subject to formal constraints
and distribution requirements.
Specifically, a recently defined class of techniques, termed {\em control
improvisation}~\cite{fremont-fsttcs15}, holds promise. An improviser is a generator of
random strings (examples) ${\bf x}$ that satisfy three constraints:
(i) a {\it hard constraint}
that defines the space of legal ${\bf x}$;
(ii) a {\it soft constraint} defining how the generated ${\bf x}$ must be similar
to real-world examples, and
(iii) a {\it randomness requirement} defining a constraint on the output
distribution. 
The theory of control improvisation is still in its infancy, and we
are just starting to understand the computational complexity and to
devise efficient algorithms. Improvisation, in turn, relies on
recent progress on computational problems such as constrained random
sampling and model counting (e.g.,~\cite{meel-beyondnp16,chakraborty-aaai14,chakraborty-tacas15}), and generative approaches based on probabilistic programming (e.g.~\cite{fremont-pldi19}). 

\noindent
{\em Quantitative Verification on the Semantic Feature Space:} 
Recall the challenge to develop techniques for verification of
quantitative requirements -- where the output of the verifier is not
just YES/NO but a numeric value.

The complexity and heterogeneity of AI-based systems means that, in general,
formal verification of specifications, Boolean or quantitative, is undecidable.
(For example, even deciding whether a state of a linear hybrid system is reachable is undecidable.)
To overcome this obstacle posed by computational complexity, one must
augment the abstraction and modeling methods discussed earlier in this section
with novel techniques for probabilistic and quantitative verification 
over the semantic feature space.
For specification formalisms that have both Boolean and quantitative semantics, in formalisms such as metric temporal logic, 
the formulation of {\em verification as optimization} is crucial to unifying computational methods from formal methods with those from the optimization literature, such as in simulation-based temporal logic falsification (e.g.~\cite{jin-tcad15,fainekos-acc15,yamaguchi-tr16}), although they must be applied to the semantic feature space for efficiency~\cite{verifai-cav19}.
Such falsification techniques can also be used for the 
systematic, adversarial generation of training data for ML components~\cite{verifai-cav19}.
Techniques for probabilistic verification, such as probabilistic model checking~\cite{prism,storm}, 
should be extended beyond traditional formalisms such as Markov chains or Markov Decision Processes to {\em verify probabilistic programs over semantic feature spaces}.
Similarly, work on SMT solving must be extended to more effectively handle
cost constraints --- in other words, {\em combining SMT solving with optimization methods} (e.g.,~\cite{shoukry-hscc17,bjorner2015nuz}).

\noindent
{\em Compositional Reasoning:}
As in all applications of formal methods, modularity will be crucial to
scalable verification of AI-based systems.
However, compositional design and analysis of AI-based systems faces some unique challenges.
First, theories of probabilistic assume-guarantee design and verification 
need to be developed for the semantic spaces for such systems, building on some promising initial work (e.g.~\cite{nuzzo-tecs19}).
Second, we suggest the use of inductive synthesis~\cite{seshia-pieee15} to generate assume-guaranteee contracts algorithmically, to reduce the specification burden and ease the use of compositional reasoning.
Third, to handle the case of components, such as perception, that
do not have precise formal specifications, we suggest techniques that 
infer component-level constraints from system-level analysis (e.g.~\cite{dreossi-nfm17}) and use such constraints to focus component-level analysis,
including adversarial analysis.

\subsection{Formal Inductive Synthesis, Safe Learning, and Run-Time Assurance}

Developing a correct-by-construction design methodology for AI-based systems, 
with associated tools, is perhaps the toughest challenge of all.
For this to be fully solved, the preceding four challenges must be successfully addressed.
However, we do not need to wait until we solve those problems in order
to start working on this one. Indeed, a methodology to ``design for verification''
may well ease the task on the other four challenges.

\noindent
{\em Formal Inductive Synthesis:} 
First consider the problem of synthesizing learning components correct by construction.
The emerging theory of {\em formal inductive synthesis}~\cite{jha-arxiv15,jha-acta17} addresses this problem.
Formal inductive synthesis is the synthesis from examples of programs that satisfy formal
specifications.
In machine learning terms, it is the synthesis of models/classifiers that additionally
satisfy a formal specification.
The most common approach to solving a formal inductive synthesis problem is
to use an {\em oracle-guided} approach. In oracle-guided synthesis, a learner
is paired with an oracle who answers queries. The set of query-response types
is defined by an {\em oracle interface}. For the example of Fig.~\ref{fig:cpsml-example}, the oracle can be a falsifier
that can generate counterexamples showing how a failure of the 
learned component violates the system-level specification. 
This approach, also known as {\em counterexample-guided
inductive synthesis}~\cite{solar-asplos06}, has proved effective in 
many scenarios. In general,
oracle-guided inductive synthesis techniques show much promise
for the synthesis of learned components by blending
expert human insight, inductive learning, and
deductive reasoning~\cite{seshia-dac12,seshia-pieee15}. 
These methods also have a close relation to the sub-field of 
{\em machine teaching}~\cite{zhu-arxiv18}.

\noindent
{\em Safe Learning by Design:} 
There has been considerable recent work on using design-time methods
to analyze or constrain learning components so as to ensure safe
operation within specified assumptions.
A prominent example is
{\em safe learning-based control} (e.g.,~\cite{akametalu-cdc14,fisac-tac18}).
In this approach, a safety envelope is pre-computed and a learning algorithm
is used to tune a controller within that envelope. 
Techniques for efficiently computing such safety envelopes based,
for example, on reachability analysis~\cite{tomlin-pieee03}, are needed.
Relatedly, several methods have been proposed for safe reinforcement
learning (see~\cite{garcia-jmlr15}).
Another promising direction is to enforce properties on ML models
through the use of semantic loss functions (e.g.~\cite{xu-icml18,dreossi-cav18}), though this problem is largely unsolved.
Finally, the use of theorem proving for ensuring correctness of 
algorithms used for training ML models (e.g.~\cite{selsam-icml17}) is also an important step towards
improving the assurance in ML based systems.

\noindent
{\em Run-Time Assurance:} 
Due to the undecidability of verification in most instances and
the challenge of environment modeling, we believe it will be difficult, if
not impossible, to synthesize correct-by-construction AI-based systems 
or to formally verify correct operation without making restrictive assumptions.
Therefore, design-time verification must be combined with {\em run-time assurance}, i.e., run-time verification and mitigation techniques. 
For example, the Simplex technique~\cite{sha2001using}
provides one approach to combining a complex, but error-prone module with a safe,
formally-verified backup module. Recent techniques for combining design-time
and run-time assurance methods (e.g.,~\cite{schierman2015runtime,desai-rv17,desai-dsn19}) have shown how unverified components, including those based on
AI and ML, can be wrapped within a runtime assurance framework to provide
guarantees of safe operation. 
However, the problems of extracting environment assumptions and 
synthesizing them into runtime monitors (e.g., as
described for introspective environment modeling~\cite{seshia-rv19}) 
and devising runtime mitigation
strategies remain a largely unsolved problem.

\section{Conclusion}
\label{sec:conclusion}

\begin{table*}
\centering
\begin{tabular}{|l|l|}
\hline
Challenges & Principles \\
\hline
\hline
Environment (incl. Human) Modeling & Active Data-Driven, Introspective, Probabilistic Modeling\\ \hline
Formal Specification & Start at System Level, Derive Component Specifications;\\ 
 & Hybrid Boolean-Quantitative Specification; Specification Mining \\ \hline
Modeling Learning Systems & Abstractions, Explanations, Semantic Feature Spaces\\ \hline
Scalable Design \& Verification & Compositional Reasoning, Controlled Randomization,\\ 
of Data and Models & Quantitative Semantic Analysis\\ \hline
Correct-by-Construction Methods & Formal Inductive Synthesis, Safe Learning by Design,\\ 
& Run-Time Assurance \\
\hline
\end{tabular}
\caption{Summary of the five challenges for Verified AI presented in this paper, and the corresponding principles proposed to address them.}
\label{tbl:summary}
\end{table*}

Taking a formal methods perspective, we have analyzed the challenge of 
developing and applying formal methods to systems
that are substantially based on artificial intelligence or machine learning.
As summarized in Table~\ref{tbl:summary}, we have identified
five main challenges for applying formal methods to AI-based systems.
For each of these five challenges, we have identified corresponding principles
for design and verification that hold promise for addressing that challenge.
Since the original version of this paper was published in 2016,
several researchers including the authors have been 
working on addressing these challenges; a few sample advances are described in this paper.
In particular, we have developed open-source tools, VerifAI~\cite{verifai-www} and 
Scenic~\cite{scenic-www} that implement techniques based
on the principles described in this paper, and which have been
applied to industrial-scale systems in the autonomous driving~\cite{fremont-itsc20}
and aerospace~\cite{fremont-cav20} domains.
These results are but a start and much more remains to be done.
The topic of Verified AI promises to continue to be a fruitful
area for research in the years to come.


\section*{Acknowledgments}

The authors' work has been supported in part by NSF grants CCF-1139138,
CCF-1116993, CNS-1545126 (VeHICaL), CNS-1646208, and CCF-1837132 (FMitF), by an NDSEG Fellowship, 
by the TerraSwarm Research Center, one of six centers supported by
the STARnet phase of the Focus Center Research Program (FCRP) a
Semiconductor Research Corporation program sponsored by MARCO and
DARPA, by the DARPA BRASS and Assured Autonomy programs, by
Toyota under the iCyPhy center, and by Berkeley Deep Drive.
We gratefully acknowledge the many colleagues with whom our conversations 
and collaborations have helped shape this article.


\bibliographystyle{plain}


\end{document}